\pdfoutput=1

\documentclass[11pt]{article}

\usepackage[]{acl}

\usepackage{times}
\usepackage{graphicx}
\usepackage{latexsym}
\usepackage{multirow}
\usepackage{booktabs}
\usepackage{todonotes}
\usepackage{amsmath}
\usepackage{tabularx,colortbl}

\usepackage{tikz-dependency}

\usepackage[T1]{fontenc}

\usepackage[utf8]{inputenc}

\usepackage{microtype}

\title{Don't Discard Fixed-Window Audio Segmentation in Speech-to-Text Translation}

\author{Chantal Amrhein$^1$ \and Barry Haddow$^{2}$\\
  $^1$Department of Computational Linguistics, University of Zurich\\
  $^2$School of Informatics, University of Edinburgh \\ \medskip
  \texttt{amrhein@cl.uzh.ch, bhaddow@ed.ac.uk}}

\begin{document}
\maketitle
\begin{abstract}
For real-life applications, it is crucial that end-to-end spoken language translation models 
perform well on continuous audio, without relying on human-supplied segmentation. 
For \emph{online} spoken language translation,  where models need to start translating before the full utterance is spoken,
most previous work has ignored the segmentation problem. In this paper, we compare various methods for improving models' robustness towards segmentation errors and different segmentation strategies in both offline and online settings and report results on translation quality, flicker and delay. Our findings on five different language pairs show that a simple fixed-window audio segmentation can perform surprisingly well given the right conditions.\footnote{We publicly release our code and model outputs here: \url{https://github.com/ZurichNLP/window_audio_segmentation}}
\end{abstract}

\section{Introduction}
\label{sec:intro}
End-to-end spoken language translation (SLT) has seen considerable advances in recent years. To apply these findings to real online and offline SLT settings, we need to be able to process continuous audio input. However, most previous work on end-to-end SLT makes use of human-annotated, sentence-like gold segments both at training and test time which are not available in real-life settings. Unfortunately, SLT models that were trained on such gold segments often suffer a noticeable quality loss when applied to artificially split audio segments \citep{zhang-etal-2021-beyond, tsiamas2022shas}. This also highlights that a good segmentation is more important for SLT than for automatic speech recognition (ASR) because we need to split the audio into ``translatable units''. For a cascade system,  a segmenter/punctuator can be inserted between the ASR and machine translation (MT) model \citep{cho2017nmt} in 
order to create suitable segments for the MT model.  However for end-to-end SLT systems,  it is still not clear how to best translate continuous input.

Solving this problem is very much an active research field that has mainly been tackled from two sides: (1) improving SLT models to be more robust towards segmentation errors \citep{Gaido2020,li2021sentence,zhang-etal-2021-beyond} and (2) developing strategies to split streaming audio into segments that resemble the training data more closely \citep{gaido-etal-2021-beyond, tsiamas2022shas}. Both types of approaches were successfully used in recent years for the IWSLT offline SLT shared task \citep{ansari-etal-2020-findings, anastasopoulos-etal-2021-findings, anastasopoulos-etal-2022-findings} to translate audio without gold segmentations. However, they have not yet been tested systematically in the online SLT setup where translation starts before the full utterance is spoken. Recent editions of the IWSLT simultaneous speech translation shared task focused more on evaluation using the gold segmentation rather than unsegmented audio  \citep{anastasopoulos-etal-2021-findings, anastasopoulos-etal-2022-findings}. Segmenting streaming audio is especially interesting in online SLT because aside from effects on translation quality, different segmentations can also influence the delay (or latency) of the generated translation.

\begin{figure}[t]
    \centering
    \includegraphics[width=0.43\textwidth]{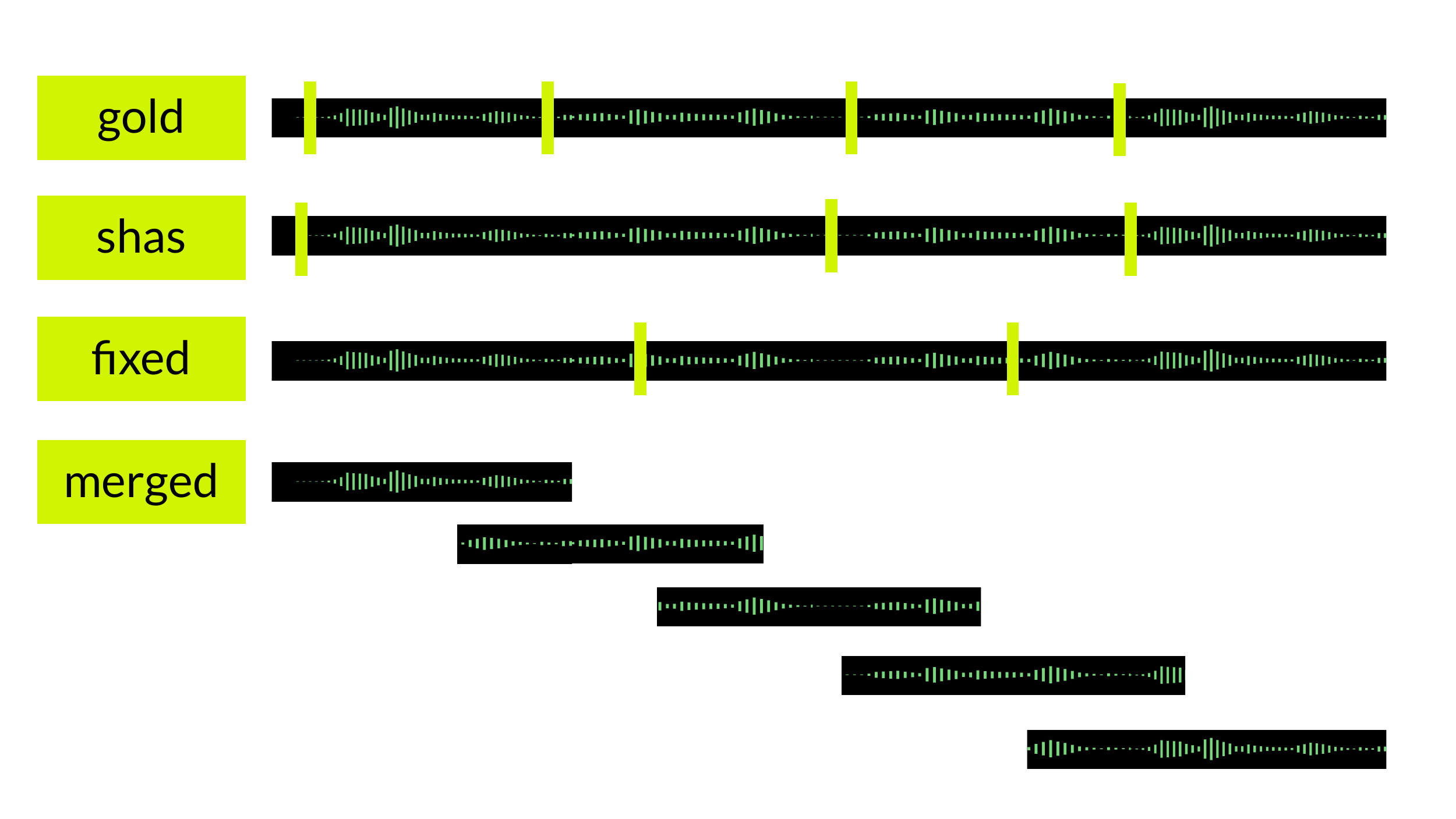}
    \caption{Visualisation of the different audio segmentation methods studied in this paper.}
    \label{fig:seg_methods}
\end{figure}

In this paper, we aim to fill this gap and focus on the end-to-end online SLT setup. We suspect that there is an interplay between more robust models and better segmentation strategies and that an isolated comparison may not be informative enough. Consequently, we explore different combinations of these two approaches for two different SLT models and present results in five language pairs. Figure \ref{fig:seg_methods} shows the four segmentation methods we study in this work (see also Section \ref{sec:segmentation}). Our experiments follow the popular retranslation approach \cite{Niehues2016DynamicTF,Niehues2018,Arivazhagan2020ReTranslationSF,arivazhagan-etal-2020-translation} where a partial segment is retranslated every time new audio becomes available. Retranslation has the advantage of being a simple approach to online SLT, which can use a standard MT inference engine.  As a side-effect, the previous translation can change in later retranslations and the resulting ``flicker'' (i.e. sudden translation changes in the output of previous time steps) is also considered in our evaluation of different strategies.\\

Our main contributions are:
\begin{itemize}
   \item We explore various combinations of segmentation strategies and robustness-finetuning approaches for translating unsegmented audio in an online SLT setup.
   \item We find that the advantage of dedicated audio segmentation models over a fixed-window approach becomes much smaller if the translation model is context-aware, and merging translations of overlapping windows can perform comparatively to the gold segmentation.
   \item We discuss issues with the evaluation of delay in an existing evaluation toolkit for retranslation when different segmentations are used and show how these can be mitigated.
\end{itemize}

\section{Related Work}

In recent years, the IWSLT shared task organisers have stopped providing gold segmented test sets for the offline speech translation task which has lead to increased research focus on audio segmentation \citep{ansari-etal-2020-findings, anastasopoulos-etal-2021-findings, anastasopoulos-etal-2022-findings}. One obvious strategy to segment audio is to create fixed windows of the same duration, but previous research has mostly relied on more elaborate methods. Typically, methods with voice activity detection (VAD) \citep{sohn1999} were employed to identify natural breaks in the speech signal. However, VAD models do not guarantee breaks that align with complete utterances and can produce segments that are too long or too short which is why hybrid approaches that also consider the length of the predicted utterance can be helpful \citep{potapczyk-przybysz-2020-srpols,gaido-etal-2021-beyond, shanbhogue-etal-2022-amazon}. Most recently, \citet{tsiamas2022shas} finetune a wav2vec 2.0 model \citep{DBLP:conf/nips/BaevskiZMA20} to predict gold segmentation-like utterance boundaries, an approach which outperforms several alternative segmentation methods and was widely adopted in the 2022 IWSLT offline SLT shared task \citep{tsiamas-etal-2022-pretrained, pham-etal-2022-effective, gaido-etal-2022-efficient}.

Apart from improving automatic audio segmentation methods, previous research has also focused on making SLT models more robust toward segmentation errors. \citet{Gaido2020} and \citet{zhang-etal-2021-beyond} both explore context-aware end-to-end SLT models and show that context can help to better translate VAD-segmented utterances. Similarly, training on artificially truncated data can be beneficial to segmentation robustness in cascaded setups \citep{li2021sentence} but also in end-to-end models \citep{Gaido2020}. While this approach can introduce misalignments between source audio and target text, such misalignments in the training data are not necessarily harmful to SLT models as \citet{ouyang-etal-2022-impact} recently showed in an evaluation of the MuST-C dataset \citep{di-gangi-etal-2019-must}.

Both of these approaches – improving automatic segmentation and making models more robust toward segmentation errors – can be combined. For example, \citet{papi-etal-2021-dealing} show that continued finetuning on artificial segmentation can help narrow the gap between hybrid segmentation approaches and manual segmentation. However, a combination of both methods is not always equally beneficial. \citet{gaido-etal-2022-efficient} repeat \citet{papi-etal-2021-dealing}'s analysis with the segmentation model proposed by \citet{tsiamas2022shas} and show that for this segmentation strategy, continued finetuning on resegmented data does not lead to an improvement in translation quality. 

In our work, we aim to extend these efforts and test various combinations of segmentation and model finetuning strategies. We are especially interested in fixed-window segmentations which have largely been ignored in SLT research but are attractive from a practical point of view because they do not require an additional model to perform segmentation. To the best of our knowledge, we are the first to perform such an extensive segmentation-focused analysis for online SLT, considering delay, flicker and translation quality for the evaluation.

\section{Experiment Setup}
\subsection{Data}
We run experiments with TED talk data in five different language pairs where the task is to translate a TED talk as an incoming stream without having any gold sentence segmentation.

For English-to-German, we use the data from the MuST-C corpus \citep{di-gangi-etal-2019-must} version 1.0\footnote{\url{https://ict.fbk.eu/must-c/}}. This dataset is built from TED talk audio with human-annotated transcriptions and translations. For testing, we use the ``tst-COMMON'' test set. For Spanish-, French-, Italian- and Portuguese-to-English, we use the data from the mTEDx corpus \citep{salesky2021mtedx}\footnote{\url{http://www.openslr.org/100}}. This dataset is also based on TED talks and provides human annotated transcriptions and translations of the audio files. For testing, we use the ``iwslt2021'' test set from the IWSLT 2021 multilingual speech translation shared task \citep{anastasopoulos-etal-2021-findings}. The dataset statistics can be seen in Table \ref{tab:data_statistics}.

\begin{table}[]
    \centering
    \small
    \begin{tabular}{ccccc}
         & \multicolumn{2}{c}{train} & \multicolumn{2}{c}{test}\\
         \cmidrule(lr){2-3} \cmidrule(lr){4-5}
         & \# talks & \# segments & \# talks & \# segments\\ 
         \cmidrule(lr){2-2} \cmidrule(lr){3-3} \cmidrule(lr){4-4} \cmidrule(lr){5-5}
        en-de & 2,043 & 229,703 & 27 & 2,641 \\\cmidrule(lr){1-5}
        es-en & \phantom{2,}378 & \phantom{2}36,263 & 15 & \phantom{2,}996 \\
        fr-en & \phantom{2,}250 & \phantom{2}30,171 & 11 & 1,041 \\
        it-en & \phantom{2,}221 & \phantom{2}24,576 & 11 & \phantom{2,}979 \\
        pt-en & \phantom{2,}279 & \phantom{2}30,855 & 11 & 1,022 \\\cmidrule(lr){1-5}
        multi & 1,128 & 121,865 & 48 & 4038\\
    \end{tabular}
    \caption{Overview of dataset statistics. The last row shows the total numbers for the multilingual model on es-en, fr-en, it-en and pt-en combined.}
    \label{tab:data_statistics}
\end{table}

\subsection{Spoken Language Translation Models}
We base all our experiments on the joint speech- and text-to-text model \citep{tang-etal-2021-fst, tang-etal-2021-improving, Tang2021AGM} released by Meta AI. For the English-German experiments, we use the model provided by \citet{tang-etal-2021-improving}\footnote{\url{https://github.com/facebookresearch/fairseq/blob/main/examples/speech_text_joint_to_text/docs/ende-mustc.md}} and for the other language pairs, we use the multilingual model provided by \citet{tang-etal-2021-fst}\footnote{\url{https://github.com/facebookresearch/fairseq/blob/main/examples/speech_text_joint_to_text/docs/iwslt2021.md}}. We refer to these models as the \textbf{original} models. These models are trained on full segments that mostly comprise one sentence:

\begin{itemize}
    \item[] \small \textbf{And like with all powerful technology, this brings huge benefits, but also some risks.}
\end{itemize}

To investigate the effects of different segmentation strategies combined with segmentation-robust models, we finetune three different variants based on each model. In each case, the finetuning data is augmented with artificially segmented data,  but no segments cross the boundaries between the individual TED talks.

\begin{itemize}
    \item \textbf{prefix:} This model is finetuned on a 50-50 mix of original segments and synthetically created prefixes (i.e. sentences where the end is arbitrarily chopped off). Finetuning on prefixes should help for translating artificially segmented audio where the segment stops in the middle of an utterance. We create prefixes of the original segments by randomly sampling a new duration for an audio segment and using the length ratio to extract the corresponding target text. An example for a prefixed version of the original segment can be seen here:
    
    \begin{itemize}
    \item[] \small \textbf{And like with all}
    \end{itemize}

    \item \textbf{context:} This model is finetuned on a mix of original segments and synthetically created longer segments. Context was already shown to help with segmentation errors by \citet{zhang-etal-2021-beyond}. This model should be able to translate segments that consist of multiple utterances. For each segment in the original 
    training set, we randomly either use the original segment (50\% of the time) or an extended segment created 
    by prepending the previous segment (25\% of the time) or the 2 previous segments (also 25\% of the time). We
    then add context-prefixed segments for each of these (possibly-extended) segments, by truncating the last concatenated segment. An example for a context-prefixed version of the original segment can be seen here:
    
    \begin{itemize}
    \item[] \small We work every day to generate those kinds of technologies, safe and useful. \textbf{ And like with all powerful technology, this brings huge benefits,}
    \end{itemize}
    
    \item \textbf{windows:} This model is finetuned on a 50-50 mix of original segments and windows of random duration. We split the audio into windows by starting at the beginning of the audio and then sampling the duration of the first window. The end of this window then becomes the start of the next window and we repeat this process until we reach the end of a TED talk. For every such window, we extract the corresponding target text from the time-aligned gold segment(s) via length ratios. This mirrors the conditions at inference time with a fixed-window segmentation where a segment can start and end anywhere in an utterance and can also comprise multiple utterances. The segment durations are sampled uniformly between 10 and 30 seconds. Note that this model will see the qualitatively poorest data out of all finetuned models because both the end of the segment and the beginning depend on length ratios which can introduce alignment errors. An example for a window version of the original segment can be seen here:
    
    \begin{itemize}
    \item[] \small or death diagnosis without the help of artificial intelligence. We work every day to generate those kinds of technologies, safe and useful. \textbf{ And like with all powerful technology, this brings huge benefits, but also some risks.} I don’t know how this debate ends, but what I’m sure of, is that the game
    \end{itemize}
\end{itemize}

All models are trained from the original checkpoint for an additional 20k steps and the last two checkpoints are averaged if more than one is saved. We do this finetuning by continuing training with the config file of the original model. For the English$\rightarrow$German MuST-C model, we train on the audio as well as the corresponding phoneme sequences based on the transcript, however, we do not use additional parallel text data during finetuning. For the multilingual mTEDx model, we only train on data for the selected language pairs and only on audio (no phoneme sequences) because this model was already finetuned on the spoken language translation task. The validation sets only contain gold segments and all models stop training due to the step limit before early stopping is triggered.

\subsection{Segmentation Strategies}
\label{sec:segmentation}
We consider four different inference-time segmentation strategies in our experiments, visualised in Figure \ref{fig:seg_methods}:

\begin{itemize}
    \item \textbf{gold:} These are human annotated segmentation boundaries that are released as part of the MuST-C and mTEDx data. This segmentation can be viewed as an oracle segmentation even though it may not necessarily be the best segmentation for all models. Using the gold segmentation in practice is unrealistic, especially in the online setting where there would be no time for a human
    to segment the audio before translation.
    \item \textbf{SHAS:} This segmentation method was recently proposed by \citet{tsiamas2022shas}. The authors finetune a pretrained wav2vec 2.0 model \citep{DBLP:conf/nips/BaevskiZMA20} on the gold segmentations and train it to predict probabilities for segmentation boundaries.  SHAS can be used both in offline and online setups using different algorithms to determine the segmentation boundaries based on the model's probabilities. Since we perform our experiments in an online setup, we use the pSTREAM algorithm to identify segments with SHAS. We set the maximum segment length to 18 seconds which the authors reported as best-performing.
    \item \textbf{fixed:} This is a simple approach that splits the audio stream into independent fixed windows of a given duration. In our experiments, we use durations of 26 seconds, which performed best in experiments by \citet{tsiamas2022shas}.
    \item \textbf{merged:} Similarly to above, we consider fixed-size windows for this segmentation strategy but here we construct overlapping windows. We use a duration of 15 seconds\footnote{We found empirically that this works better than a duration of 26 seconds as for fixed-windows, with both increased translation quality and reduced flicker (see Appendix \ref{app:last-common}).} and shift the window with a stride of 2 seconds at a time. The translations of these overlapping windows are merged before the next window is translated (see Section \ref{sec:algo}). 
\end{itemize}

\subsection{Retranslation}
We employ a retranslation strategy \citep{Niehues2016DynamicTF,Niehues2018,Arivazhagan2020ReTranslationSF,arivazhagan-etal-2020-translation} for our end-to-end SLT experiments.
This means that we retranslate the incoming audio at fixed time intervals. In our experiments, we retranslate every 2 seconds
to be consistent with the 2-second stride from the merging windows approach.
Because of such retranslations of the full audio segment — from the start of the segment up to the current time step — the SLT model may correct translation mistakes from earlier time steps. This means that the final translation of a complete segment reaches the quality of offline translation. However, if these updated partial translations are presented to users and there are changes to previously translated text, this may be hard to follow. Therefore, it is important to not only evaluate the quality of the translations and the delay but also how often previously translated words are changed which is termed ``flicker''. Typically, when delay improves there will be more flicker because translating sooner means a higher chance of errors that need to be corrected in the next retranslation.

\subsection{Window Merging Algorithm}
\label{sec:algo}
One reason why a fixed-window segmentation might underperform compared to other segmentations is that utterances are likely to be split up into two or multiple segments which can introduce ambiguities and result in disfluent translations. However, this problem can be reduced if the windows are overlapping which is technically very easy to do. With a retranslation approach, we can simply shift the whole window by X seconds to obtain overlapping translations. 

To merge the resulting translations, we employ a merging algorithm that was previously proposed for a cascaded SLT setup \citep{anonymous2022}. Their merging window algorithm also works for end-to-end SLT because it is not dependent on a transcript of the source audio. The algorithm identifies the longest common substring (LCS) between the growing translation of the output stream and the translation of the current window. The current output is formed by everything to the left of the LCS coming from the output at the previous time step, followed by the LCS and then everything to the right of the LCS from the current translation output. In this way, the translation of the input stream is continuously extended.

The merging is controlled by a threshold that defines the minimum required length of the LCS. At every time step, this threshold is computed by:

\begin{align*}
threshold = |T_t| * \tau 
\end{align*}

Where $T_t$ is the current window translation length and $\tau$ is a ratio hyperparameter. If the LCS is shorter than this minimum length, instead of merging the current translation with the output stream, the window is backtracked to the left and a longer window is translated. We backtrack 0.1 seconds at a time for a maximum of three backtracks. Only when a sufficiently long LCS is found or the maximum number of backtracks is reached, do we perform the merging operation. In our experiments, we set the ratio $\tau$ to 0.4 which performed best in the cascaded setup \citep{anonymous2022}. If there are multiple LCS (common substrings with the same length), we merge at the last-occurring one.

\subsection{Evaluation}
\label{sec:eval}
For evaluation, we use SLTev\footnote{\url{https://github.com/ELITR/SLTev}} \citep{ansari-etal-2021-sltev}, a toolkit that can evaluate translation quality, delay and flicker in a retranslation SLT setup. We explain below how the evaluation is 
adapted for unsegmented input. Since we assume our input is segmented at the talk level, we evaluate at the talk level
too.

For \textbf{translation quality}, SLTev internally resegments the translations and aligns the new segments to the reference segments such that the word error rate is minimised \citep{matusov-etal-2005-evaluating}. It is not guaranteed that the new segments follow the sentence boundaries and are perfectly aligned but, as long as the introduced alignment errors are similar for different segmentations, they can be compared.

For \textbf{flicker}, we cannot use the sentence-level measure in SLTev because this is computed as an average over all segment-level flicker scores, and with different segmentations, this measure is not comparable. However, the document-level measure is evaluated independent of the segmentation and this works well for our purpose.

For \textbf{delay}, we do not use the official implementation in SLTev because of the way it assigns timestamps 
to repeated tokens.
To explain the problem, consider the following example:

\begin{center}
\vspace{0.2cm}
\small
\begin{tabular}{ccl}
P & 13.18 & O \\
P & 14.18 & O horror,\\
P & 15.18 & O horror, terror, horror\\
C & 16.18 & O horror, horror, horror.
\vspace{0.2cm}
\end{tabular}
\end{center}

where we retranslate the newly available audio every second and consequently get three partial translations (P) and one final, complete translation (C). In SLTev, every token is assigned the time stamp of its type's first occurrence. This results in the following time stamp assignments with the original implementation.

\begin{dependency}[theme = simple]
   \begin{deptext}[column sep=0pt, font=\small]
      O \& horror \& , \& horror \& , \& horror \& .\\
      \\
      13.18 \& 14.18 \& 14.18 \& 14.18 \& 14.18 \& 14.18 \& 16.18\\
   \end{deptext}
    \begin{scope}
    \draw (\wordref{1}{1}.south)--(\wordref{3}{1}.north);
    \draw (\wordref{1}{2}.south)--(\wordref{3}{2}.north);
    \draw (\wordref{1}{3}.south)--(\wordref{3}{3}.north);
    \draw (\wordref{1}{4}.south)--(\wordref{3}{4}.north);
    \draw (\wordref{1}{5}.south)--(\wordref{3}{5}.north);
    \draw (\wordref{1}{6}.south)--(\wordref{3}{6}.north);
    \draw (\wordref{1}{7}.south)--(\wordref{3}{7}.north);
  \end{scope}
\end{dependency}

All occurrences of ``horror'' and ``,'' are assigned the timestamp 14.18 even though most of them are not yet generated by that time. If we translate longer segments that may be comprised of multiple sentences, encountering tokens that were already seen before becomes more and more likely. All of those would be assigned the timestamp of the first occurrence which favours longer segments (which we take to the extreme with our merged windows output stream). To solve this issue, we adapt the delay computation and store the individual timestamps for all repeated tokens. For this, we also need to be aware that previous content can change with each retranslation (e.g. terror $\rightarrow$ horror). We solve this following \citet{arivazhagan-etal-2020-translation}'s notation of content delay and only assign timestamps once the previous context has finalised:

\begin{dependency}[theme = simple]
   \begin{deptext}[column sep=0pt, font=\small]
      O \& horror \& , \& horror \& , \& horror \& .\\
      \\
      13.18 \& 14.18 \& 14.18 \& 16.18 \& 16.18 \& 16.18 \& 16.18\\
   \end{deptext}
    \begin{scope}
    \draw (\wordref{1}{1}.south)--(\wordref{3}{1}.north);
    \draw (\wordref{1}{2}.south)--(\wordref{3}{2}.north);
    \draw (\wordref{1}{3}.south)--(\wordref{3}{3}.north);
    \draw (\wordref{1}{4}.south)--(\wordref{3}{4}.north);
    \draw (\wordref{1}{5}.south)--(\wordref{3}{5}.north);
    \draw (\wordref{1}{6}.south)--(\wordref{3}{6}.north);
    \draw (\wordref{1}{7}.south)--(\wordref{3}{7}.north);
  \end{scope}
\end{dependency}

With these new timestamps, all possible segmentations will receive the same delay if the translated text is identical and longer segments are no longer favoured in the SLTev delay calculation. However, since we wait until the context has finalised before we assign the time stamps, the new delay measure is now also affected by flicker.
\begin{table}[ht!]
    \centering
    {\setlength{\tabcolsep}{5pt}
    \begin{tabular}{lccccc}
    & & {\small original} & {\small prefix} & {\small context} & {\small window} \\
      \cmidrule(lr){3-3} \cmidrule(lr){4-4} \cmidrule(lr){5-5} \cmidrule(lr){6-6} \\
    \multirow{5}{*}{en-de} & gold   & 25.4  & 25.5 & 25.2 & 25.5 \\ \cmidrule(lr){2-6}
    & SHAS   & 24.5 & 23.9 & 24.9 & 24.8\\
    & fixed  & 22.4 & 21.1 & 23.6 & 23.1\\
    & merged & 24.8 & 23.8 & \colorbox[HTML]{B2EAB1}{\textbf{25.3}} & 22.8\\\\			
    \multirow{5}{*}{es-en} & gold   & 41.6 & 41.3 & 41.1 & 41.4\\ \cmidrule(lr){2-6}
    & SHAS   & 40.2 & 40.3 & 40.7 & 41.0\\
    & fixed  & 35.0 & 36.9 & 39.6 & 38.4 \\
    & merged & 38.9 & 39.9 & \colorbox[HTML]{B2EAB1}{\textbf{42.0}} & 39.7\\\\			
    
    \multirow{5}{*}{fr-en} & gold   & 37.2 & 36.2 & 35.6 & 35.6\\ \cmidrule(lr){2-6}
    & SHAS   & \colorbox[HTML]{B2EAB1}{\textbf{36.2}} & 36.1 & 35.8 & 36.1\\
    & fixed  & 31.0 & 32.0 & 34.5 & 32.9 \\
    & merged & 34.6 & 35.2 & 35.8 & 31.9 \\\\			
    
    \multirow{5}{*}{it-en} & gold  & 27.0 & 28.7 & 28.8 & 29.0 \\ \cmidrule(lr){2-6}
    & SHAS   & 26.4 & 28.0 & 28.7 & 29.0 \\
    & fixed  & 22.5 & 25.6 & 27.5 & 26.3 \\
    & merged & 25.3 & 27.4 & \colorbox[HTML]{B2EAB1}{\textbf{29.2}} & 27.6\\\\
    
   \multirow{5}{*}{pt-en} & gold  & 30.6 & 29.5 & 28.7 & 29.1\\ \cmidrule(lr){2-6}
    & SHAS  & \colorbox[HTML]{B2EAB1}{\textbf{29.5}} & 28.9 & 29.2 & 28.6\\
    & fixed  & 23.6 & 24.0 & 26.9 & 26.2 \\
    & merged & 26.6 & 27.4 & 28.1 & 24.3\\\\  			
    \end{tabular}
    \caption{BLEU scores with different SLT models (columns) and different audio segmentation methods (rows). Best result for \textit{automatic} segmentation scenario marked in bold and green.}
    \label{tab:bleu}
    }
\end{table}

\section{Results}
\label{sec:results}
\subsection{Translation Quality}
We compare the different SLT models on different segmentations of the test sets and show the resulting translation quality of the complete segments in terms of BLEU in Table \ref{tab:bleu}. Note that we would reach the same translation quality in an offline setting because the final retranslation is a translation of the full window, and in common with previous work, translation quality of online SLT is only measured on the final retranslation. We also evaluate with COMET \citep{rei-etal-2020-comet} and report even better results with the merging windows approach but also find that COMET might be less reliable in a streaming SLT setup due to resegmentation errors (see Section \ref{sec:comet}).

\textbf{Does SHAS perform best with the original model (first column) as in previous work?} When the SLT model is just trained on gold data, SHAS proves to be the best-performing segmentation out of all automatic segmentations which is in line with results by \citet{tsiamas2022shas} and \citet{gaido-etal-2022-efficient}. As in previous studies, we also find that the original model shows a considerable drop in BLEU when moving from the gold segmentation to automatically split segments.

\textbf{Is SHAS still the best-performing segmentation with the finetuned models?} 
Finetuning with alternative segmentations can offer strong improvements for SHAS (+2.3) on it-en, with small improvements on 
es-en and en-de, but lower BLEU on pt-en and fr-en.
Similarly, \citet{gaido-etal-2022-efficient} found that SHAS did not benefit from finetuning on resegmented data.
However, for the two segmentation approaches based on fixed windows, finetuning greatly reduces the gap to the gold segmentation.

This is especially noticeable when we finetune on context and prefixes (third column). This confirms the finding by \citet{zhang-etal-2021-beyond} that context-aware models can better translate artificially segmented audio. When merging overlapping windows, we consistently see an improvement over the segmentation with non-overlapping fixed windows. In three language pairs, this method outperforms SHAS and in two the context-finetuned model even improves over the gold segmentation.

\begin{figure*}[ht]
    \centering
    \includegraphics[width=0.8\textwidth]{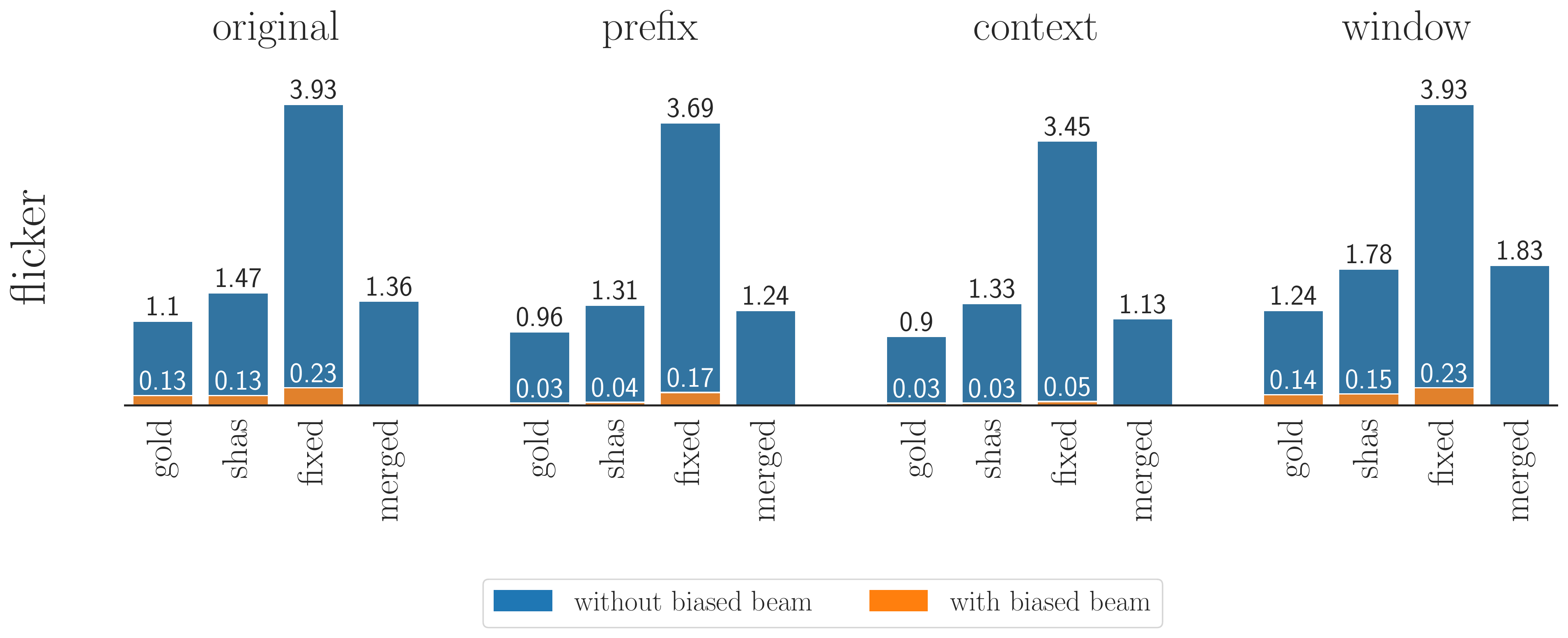}
    \caption{Flicker values for the different segmentation strategies and SLT models on the Spanish-to-English test set. The results are grouped by training strategy and each bar corresponds to a different segmentation strategy. We do not apply biased beam search to the merged segmentation.}
    \label{fig:flicker}
\end{figure*}

\textbf{Do training conditions need to match the segmentation at inference time?}
Apart from the context-aware finetuned model, we also finetuned a model on fixed windows of random duration (last column). This matches the fixed-window audio input at inference time better because a segment can start anywhere in an utterance, unlike the context-based model where every training segment started at the beginning of an utterance. Surprisingly, we find that the model finetuned on windows of random duration generally performs worse with the merging window strategy than the context-based model. This suggests that the training data for this model contains more misalignments between speech and translation because we extract both the start and the end of the segment via length ratios. This causes more flicker (see next Section) which makes it harder to merge the translations at each time step correctly. We leave extended experiments where the alignments between speech and translation are computed via ASR or the SLT output of the windows of random durations (as opposed to a simple length ratio) to future work.

\begin{figure*}[t]
    \centering
    \includegraphics[width=0.8\textwidth]{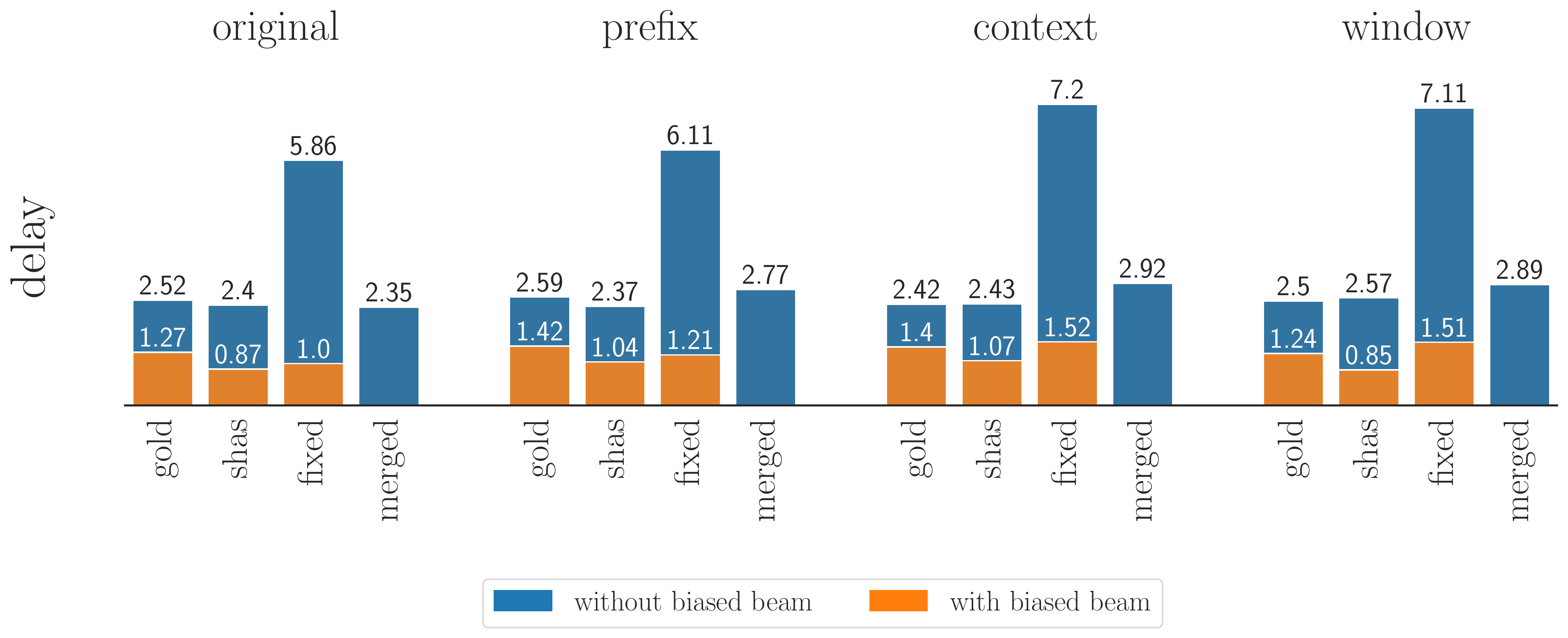}
    \caption{Delay values for the different segmentation strategies and SLT models on the Spanish-to-English test set. The results are grouped by training strategy and each bar corresponds to a different segmentation strategy.}
    \label{fig:delay}
\end{figure*}

\subsection{Flicker}
\label{subsec:flicker}

As mentioned in the \hyperref[sec:intro]{Introduction}, translation quality is not the only important evaluation metric in an online SLT scenario. When using a retranslation approach, we also need to consider the flicker that is caused by the model updating its translations at every time step. We compute the flicker as described in Section \ref{sec:eval}. The flicker scores for the Spanish-to-English test set can be seen in Figure \ref{fig:flicker}, the same figures for the other language pairs are in Appendix \ref{app:other-results}. For the results shown here, we use an output mask of 0. We show in Appendix \ref{app:output-mask} that our findings also hold with larger output masks. We show scores with and without biased beam search. 

Biased beam search \citep{Arivazhagan2020ReTranslationSF} is a modification to regular beam search that biases the probability distribution at the current time step towards a token in a given prefix translation at the same timestep. This 
can be used to stabilise retranslation -- the translation of the current prefix is biased towards the translation of the previous
prefix, suppressing flicker. In our experiments, we use the translation of the previous step as the prefix with a beta parameter of 0.25 and mask the 5 last tokens such that changes towards the end of the sentence are still possible\footnote{We do not show translation quality scores with biased beam search because on average there is only a difference of -0.006 BLEU.}. Biased beam search cannot be applied directly to the merged window approach, since it depends
on an alignment between the translation of the current prefix and that of the previous prefix. When translating using sliding windows,
the current and previous prefixes have different start points, so their translations cannot be easily aligned. We experimented with a way to reduce flicker by merging on the last common substring rather than the longest but this causes considerable translation quality loss (see Appendix \ref{app:last-common}).

\textbf{Does the segmentation strategy matter for flicker?} From Figure \ref{fig:flicker}, we can see that there are big differences between the different segmentation strategies. Fixed windows have the highest flicker because there we translate the longest windows. If something at the beginning of the window translation is changed, this will increase the flicker score considerably. With biased beam search, the flicker can be dramatically reduced. Merging overlapping windows has a lower flicker than fixed windows without biased beam search, both because the duration of the windows is shorter and because the merging algorithm prohibits changes to the left of the longest common substring\footnote{Reducing the window length to 15 seconds for the fixed window segmentation reaches a flicker that is only slightly higher than for merged windows but the translation quality suffers considerably.}. This segmentation method even has lower flicker than SHAS when no biased beam search is applied. With biased beam search, SHAS performs mostly similar to the gold segmentation which has the lowest flicker overall.

\textbf{Does model finetuning help reduce flicker?} Prefix finetuning helps reduce flicker both with and without biased beam search because the models see incomplete sentences at training time and are less likely to hallucinate to finish the sentence. 
Context finetuning helps even more and we saw in the outputs that this model has less of a tendency to connect multiple sentences into a longer sentence which can reduce flicker. The model finetuned on windows shows an even higher flicker than the original model for most segmentation strategies even though it was designed to be able to translate segments that can start and end anywhere in a sentence. As discussed in the previous section, we think this increased flicker is an artefact of the automatically generated training data which can be erroneous.

\subsection{Delay}
\label{subsec:delay}
The final evaluation metric we consider is delay. The results can be seen in Figure \ref{fig:delay}. Again, we show results with and without biased beam search for the gold, SHAS and fixed-window segmentation.

\textbf{Does the segmentation strategy matter for delay?} Because our definition of delay is affected by flicker as well (see Section \ref{sec:eval}), the fixed segmentation without biased beam search not only has the highest flicker but also the highest delay. In our results, we can see that the high delay is caused by the flicker because when we reduce flicker with biased beam search the fixed segmentation has comparable delay to the gold and SHAS segmentations. The merging windows approach has comparable delay to the gold and SHAS segmentations without biased beam search. Since we cannot apply biased beam search reliably to the merging windows approach without hurting translation quality, the flicker cannot be reduced and therefore, the merging windows approach has higher delay than the other segmentation methods with biased beam search. If delay could be defined independently of flicker in a way that still works for comparing different segmentations, the merging windows approach would likely have similar delay also compared to the outputs with biased beam search.

\textbf{Does model finetuning help reduce delay?} The results are a bit mixed. For example, the context model reduces delay for the gold segmentation but increases it slightly for SHAS and more for the fixed segmentation and the merging windows approach. In general, the choice of the model does not seem to be as important for delay as for translation quality and to a lesser extent flicker. It is possible that apparent effects only occur because our definition of delay is affected by flicker.

\section{Discussion}
Based on our results in Section \ref{sec:results}, we believe that fixed-window segmentation should not be disregarded in future SLT research on unsegmented audio. Given the right setup with a context-aware model and a merging window algorithm, this segmentation can outperform current state-of-the-art automatic segmentation models and in some cases even the gold segmentation in terms of translation quality. Moreover, in an online SLT setup, a fixed-window approach brings the additional benefit that no dedicated segmentation model needs to be loaded at inference time and run every time new audio becomes available.

While there is currently no solution to bring flicker down to biased beam search levels without hurting quality (see Appendix \ref{app:last-common}) or increasing delay (see Appendix \ref{app:output-mask}), this should not be a reason to disregard fixed-window segmentation as it opens exciting opportunities for future research. 

\section{Conclusion}
In this paper, we explored several combinations of segmentation-robust finetuning and different automatic segmentation strategies in an online SLT setup. We focus on a retranslation-based approach to SLT and we run experiments on five different language pairs based on two different SLT models. Considering the evaluation of translation quality, flicker and delay, we discuss several issues that arise when comparing different segmentations and propose a fix to an existing toolkit for evaluating delay. Our results show that a simple fixed-window segmentation can perform surprisingly well if an algorithm is used for merging overlapping windows and a context-aware SLT model is used. In terms of translation quality, this segmentation performs comparably to SHAS — the current state-of-the-art segmentation method — and in some cases even outperforms the gold segmentation, showing potential for future application to offline SLT. In terms of flicker and delay, the results of the merging windows approach are comparable to the other segmentations if biased beam search is not enabled but future work is needed to reduce flicker in the merging windows approach to similar levels as biased beam search for other strategies without hurting translation quality. 

\section*{Ethical Considerations}
In our work, we only use publicly available model checkpoints, toolkits and datasets and do not collect any additional data. Our experiments also do not involve human annotators.

\section*{Limitations}
While we aim to evaluate on a number of language pairs and with different automatic metrics, there are still some open questions that we could not answer in this work. First, we did not perform a human evaluation and, therefore, it remains unclear how distracting the different flicker and delay values with different setups would be for a user. However, previous work by \citet{machacek2020mtedx} shows that character erasure - a metric related to flicker - correlates with usability scores in a human evaluation which suggests that this would also be true for flicker. Second, the current implementation of SHAS can be used to simulate an online setting but it still expects the full audio as input. Consequently, we could not empirically compare how long translation takes with different segmentation methods in a real online setup.
Third, our experiments are limited to  SLT using a retranslation strategy. We leave further experiments with simultaneous SLT models that use a policy to decide at each time step whether to wait for further input or to translate for the future.

\section*{Acknowledgements}
We thank Sukanta Sen and Ioannis Tsiamas who shared their code with us for the window merging algorithm and the pSTREAM implementation of SHAS, respectively. We are also grateful to Noëmi Aepli and Amit Moryossef and the anonymous reviewers for their valuable feedback. This work has received funding from the European Union’s Horizon 2020 Research and Innovation Programme under Grant
Agreements (project ELITR; no. 825460 and project GoURMET; no. 825299), and from the Swiss National Science Foundation (project MUTAMUR; no. 176727). All experiments made use of the infrastructure services provided by S3IT, the Service and Support for Science IT team at the University of Zurich.

\bibliography{anthology,custom}
\bibliographystyle{acl_natbib}

\appendix
\newpage
\section*{Appendix}

\section{Further Finetuning Specifications}
We finetune all models and translate with a single NVIDIA Tesla V100 GPU. For the multilingual mTEDx model, the additional parameter \texttt{load-speech-only} needs to be added to the official training script\footnote{\url{https://github.com/facebookresearch/fairseq/blob/main/examples/speech_text_joint_to_text/docs/iwslt2021.md}}. We use the \texttt{restore-file} parameter to specify the checkpoints of the original models from which continued training should be initialised.

We will release all code (training scripts, translation scripts and evaluation modifications), the finetuned model checkpoints and the outputs upon publication.

\section{Experiments with Last Common Subsequence}
\label{app:last-common}

\begin{figure}[h]
    \centering
    \includegraphics[width=0.45\textwidth]{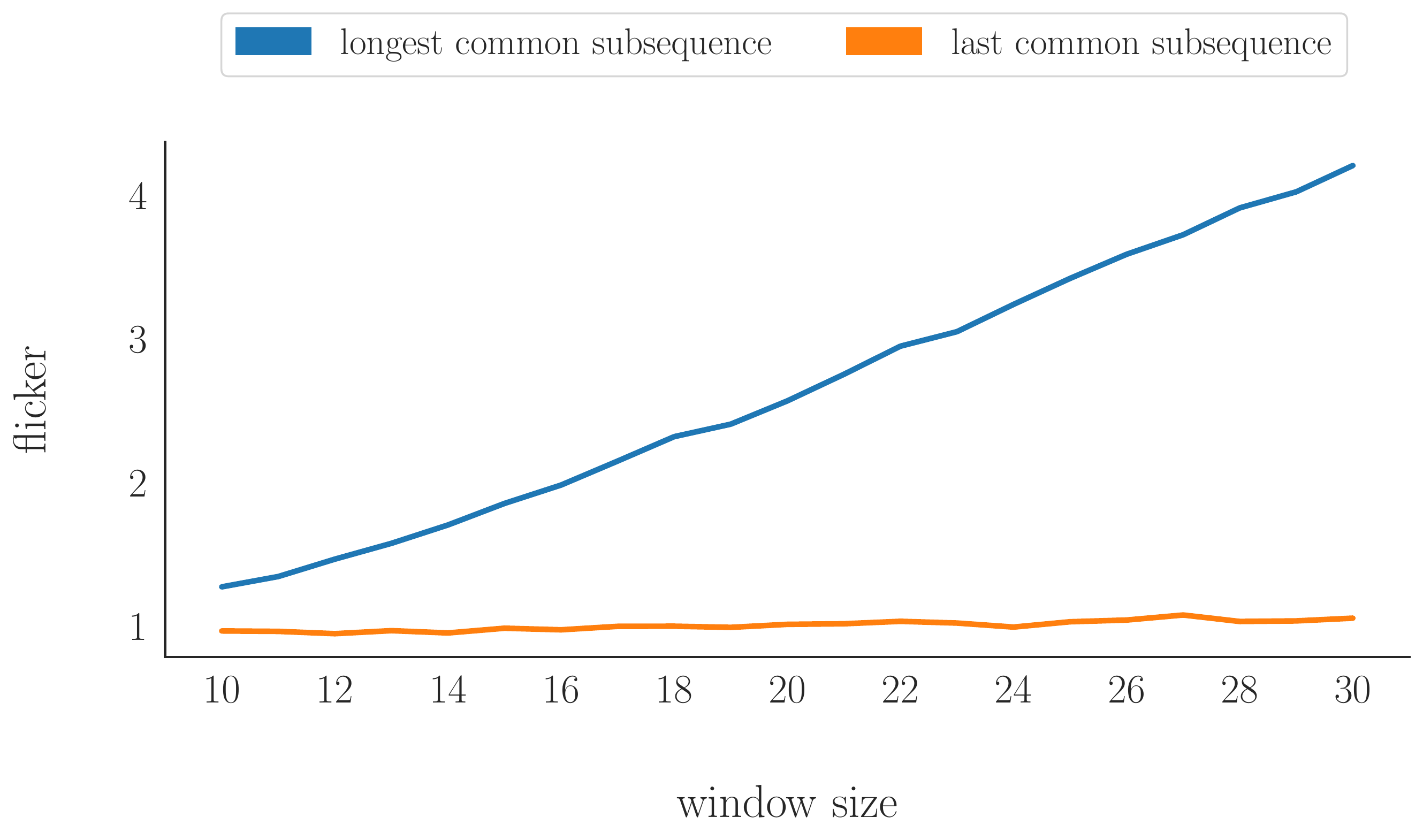}
    \caption{Flicker values with the original model on the English-to-German test set for different window sizes when merging on the longest common subsequence (blue) and the last common subsequence (orange).}
    \label{fig:last_longest_flicker}
\end{figure}

As a possible way of reducing flicker for the merging windows approach, we try merging on the last common subsequence (longer than two tokens) instead of the longest common subsequence. In this way, we can maximise the finalised part of the growing output translation and reduce flicker. Figure \ref{fig:last_longest_flicker} shows how the flicker increases for both merging strategies when the window size increases. With the original implementation that merges on the longest common subsequence, the flicker increases dramatically when the window size is increased. For the modified merging algorithm that merges on the last subsequence (longer than two tokens) the flicker increases only moderately with increased window size and is in general much lower. 

\begin{figure}[h]
    \centering
    \includegraphics[width=0.45\textwidth]{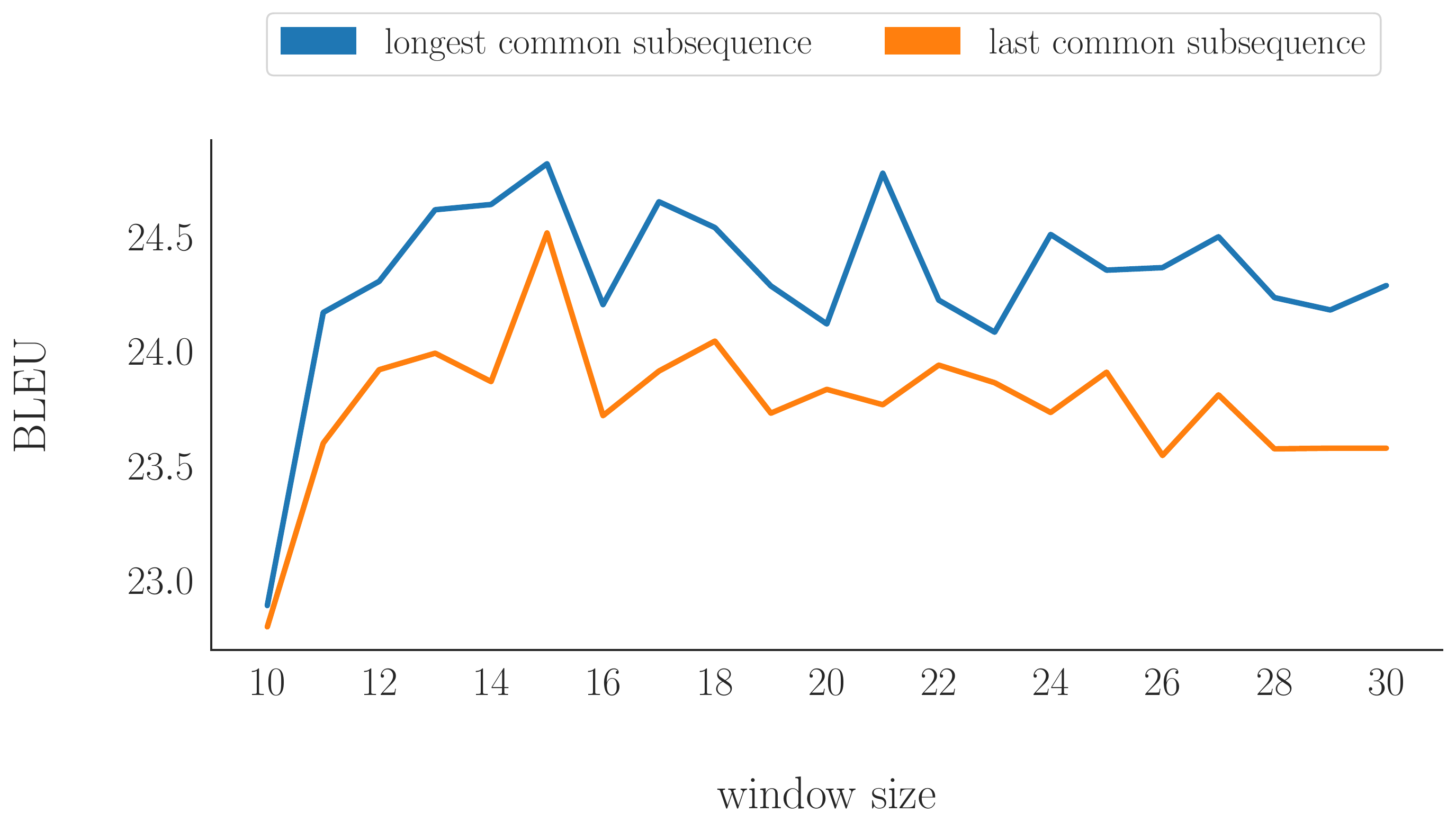}
    \caption{BLEU scores with the original model on the English-to-German test set for different window sizes when merging on the longest common subsequence (blue) and the last common subsequence (orange).}
    \label{fig:last_longest_quality}
\end{figure}

Based on these results, one might choose to merge on the last sequence, however, this change also affects the translation quality. Figure \ref{fig:last_longest_quality} shows the BLEU scores of both merging methods with different window sizes. Unfortunately, merging on the last common subsequence performs continuously worse than merging on the longest common subsequence. If quality is the main focus, this merging method is not advisable. These results also show that a window size of 15 performs best for the merging windows approach.

\section{Results with Output Mask}
\label{app:output-mask}

We also evaluate the four different segmentation methods when an output mask is applied. This means at every time step the output is truncated from the right. The number of tokens that are removed is defined by the mask size, i.e. a mask of size 0 means no tokens are removed and a mask of size 7 means seven tokens are removed. We compute these results for Spanish-to-English without biased beam search and the context-aware model which showed the lowest flicker in general. 

\begin{figure}[h]
    \centering
    \includegraphics[width=0.45\textwidth]{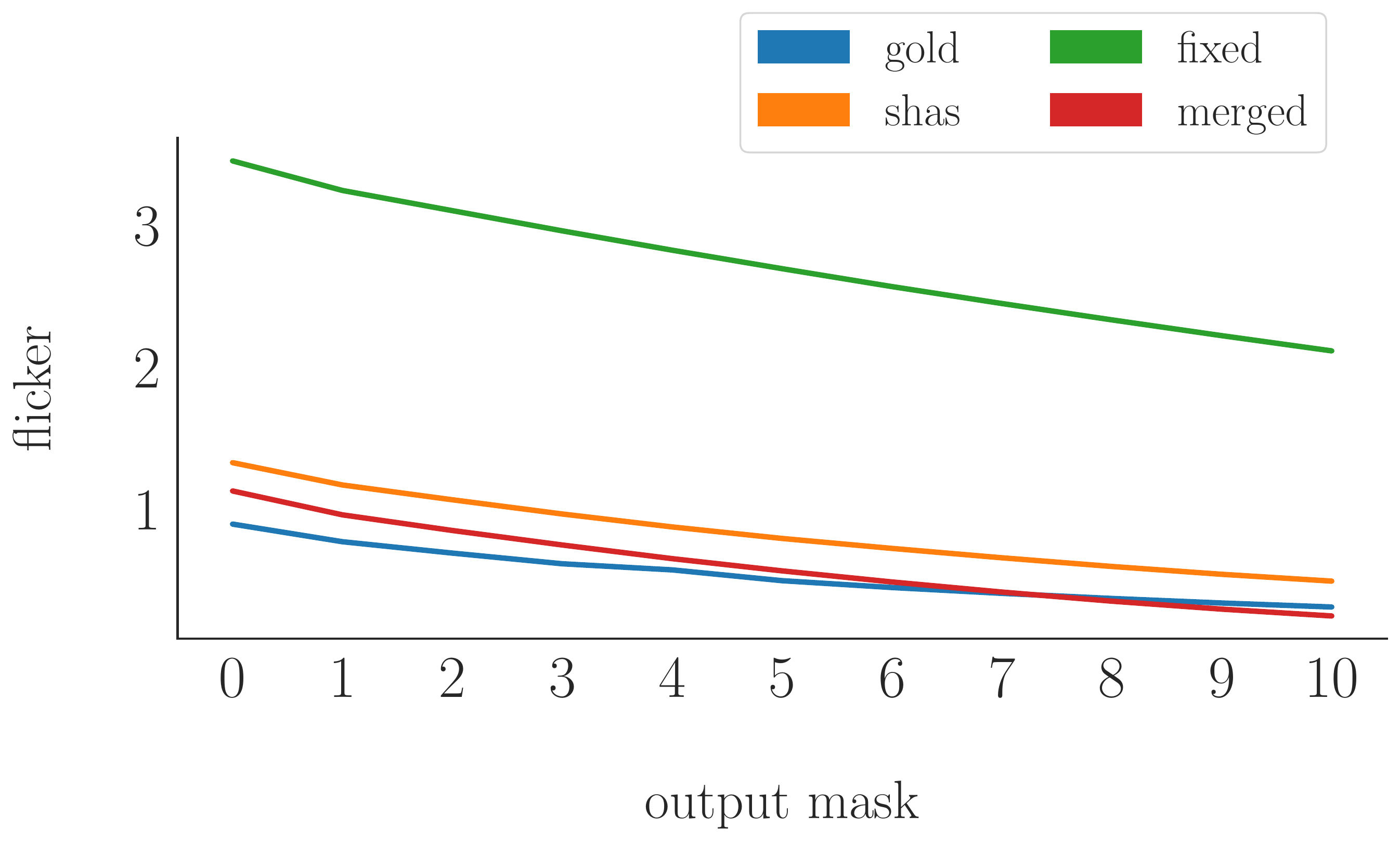}
    \caption{Flicker with different output masks on the Spanish-to-English test set. Results for all four segmentation methods with the context-finetuned model.}
    \label{fig:output_mask_flicker}
\end{figure}

Figure \ref{fig:output_mask_flicker} shows the flicker at different output mask sizes. First of all, it can be noticed that the fixed window segmentation has a continuously higher flicker than all other segmentation methods and that the flicker is still rather large even with a mask of size 10. This suggests that most flicker in the fixed-window segmentation does not occur towards the end of the segments.

The merging windows approach consistently has lower flicker than SHAS and with larger mask sizes even lower flicker than the gold segmentation. With a mask of size 10, the flicker is at 0.25 which is comparable to the flicker of the original model with fixed window segmentation where biased beam search is enabled.

\begin{figure}[h]
    \centering
    \includegraphics[width=0.45\textwidth]{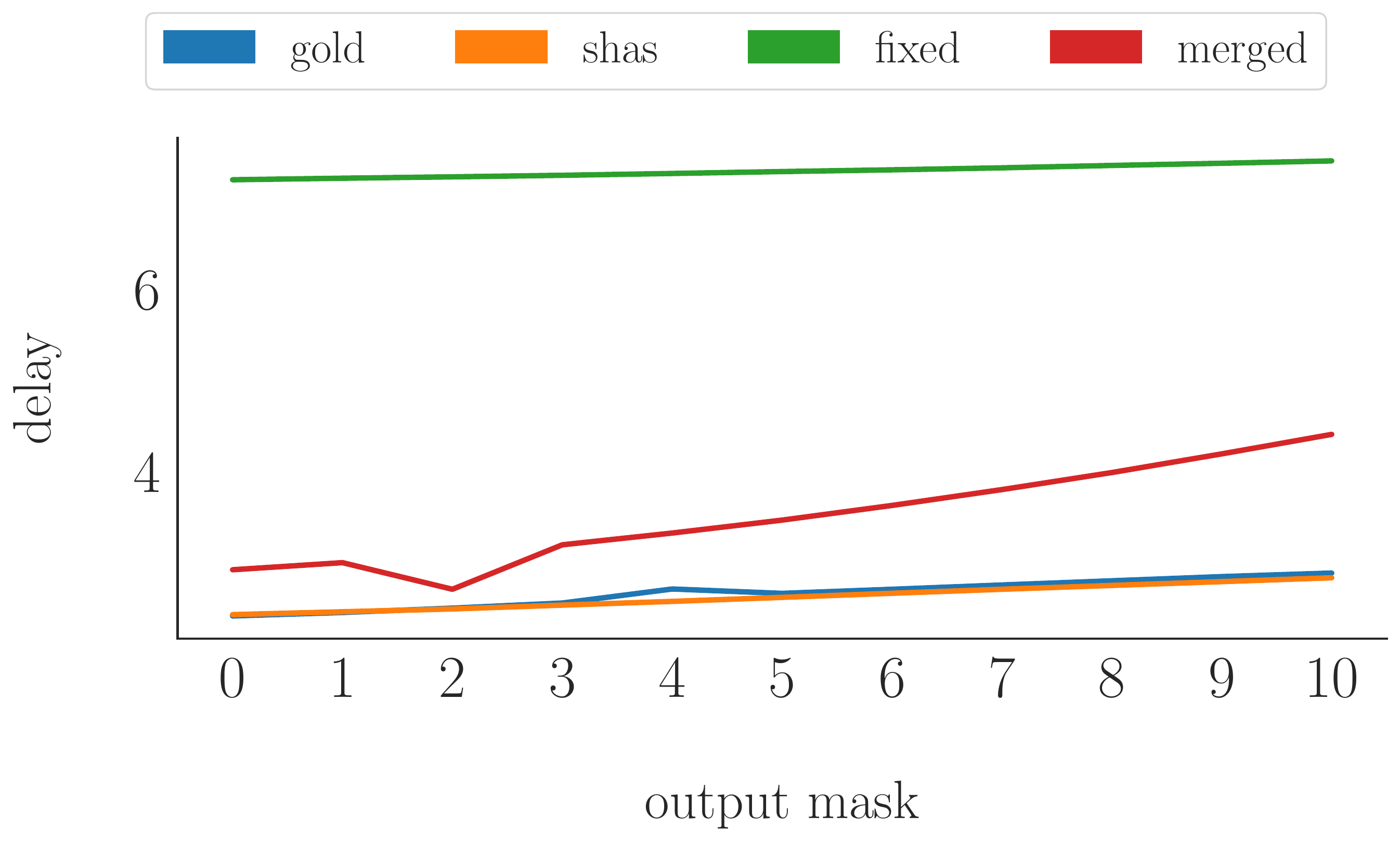}
    \caption{Delay with different output masks on the Spanish-to-English test set. Results for all four segmentation methods with the context-finetuned model.}
    \label{fig:output_mask_delay}
\end{figure}

However, this reduced flicker comes at the cost of a higher delay because the masked tokens will not be available at the time they are actually produced. This flicker-delay trade-off is well-known. Figure \ref{fig:output_mask_delay} shows the increase in delay with larger output masks. For the merging windows approach, we see that the delay increases more than for SHAS and the gold segmentation. Since our definition of the delay measure is affected by flicker, these results are hard to interpret. Nevertheless, using an output mask is a way to reduce flicker for the merging windows approach without reducing translation quality but we need to accept a higher delay.

\section{Additional Results}
\label{app:other-results}

\begin{table}[ht!]
    \centering
    \small
    {\setlength{\tabcolsep}{5pt}
    \begin{tabular}{lccccc}
    & & {\small original} & {\small prefix} & {\small context} & {\small window} \\
      \cmidrule(lr){3-3} \cmidrule(lr){4-4} \cmidrule(lr){5-5} \cmidrule(lr){6-6} \\
    \multirow{5}{*}{en-de} & gold   & -0.0589  & -0.0801 & -0.0659 & -0.0696 \\ \cmidrule(lr){2-6}
    & SHAS   & -0.1762 & -0.1934 & -0.0835 & -0.1418 \\
    & fixed  & -0.3080 & -0.3655 & -0.1846 & -0.1671 \\
    & merged & -0.1821 & -0.2169 & \colorbox[HTML]{B2EAB1}{\textbf{-0.0683}} & -0.1133\\\\			
    \multirow{5}{*}{es-en} & gold & \phantom{-}0.3175 & \phantom{-}0.2864 & \phantom{-}0.2776 & \phantom{-}0.2981 \\ \cmidrule(lr){2-6}
    & SHAS   & \phantom{-}0.2145 & \phantom{-}0.2291 & \phantom{-}0.2784 & \phantom{-}0.2638 \\
    & fixed  & -0.0339 & \phantom{-}0.0448 & \phantom{-}0.2637 & \phantom{-}0.2633 \\
    & merged & \phantom{-}0.2658 & \phantom{-}0.2736 & \colorbox[HTML]{B2EAB1}{\textbf{\phantom{-}0.3962}} & \phantom{-}0.3642\\\\			
    
    \multirow{5}{*}{fr-en} & gold   & \phantom{-}0.1702 & \phantom{-}0.1380 &  \phantom{-}0.1123 & \phantom{-}0.1078\\ \cmidrule(lr){2-6}
    & SHAS   & \phantom{-}0.1147 & \phantom{-}0.1421 & \phantom{-}0.1742 & \phantom{-}0.134\\
    & fixed  & -0.1696 & -0.1115 & \phantom{-}0.0777 & \phantom{-}0.0316\\
    & merged & \phantom{-}0.0978 & \phantom{-}0.1066 & \colorbox[HTML]{B2EAB1}{\textbf{\phantom{-}0.2170}} & \phantom{-}0.1109\\\\			
    
    \multirow{5}{*}{it-en} & gold  &\phantom{-}0.0566  &  \phantom{-}0.0583&  \phantom{-}0.0704& \phantom{-}0.0886\\ \cmidrule(lr){2-6}
    & SHAS   &-0.012 & \phantom{-}0.0215&\phantom{-}0.0915 & \phantom{-}0.0709\\
    & fixed  & -0.305& -0.2072& \phantom{-}0.0142 & -0.0408\\
    & merged & -0.0536 & -0.0066 & \colorbox[HTML]{B2EAB1}{\textbf{\phantom{-}0.1255}} & \phantom{-}0.0619 \\\\
    
   \multirow{5}{*}{pt-en} & gold  & \phantom{-}0.0662 & \phantom{-}0.0234 & -0.0130 & -0.0048\\ \cmidrule(lr){2-6}
    & SHAS  & -0.0108 & -0.0104 & -0.0085 & -0.0085\\
    & fixed  & -0.2939 & -0.2853& -0.0854 & -0.0937 \\
    & merged & -0.0581 & -0.0784 & \colorbox[HTML]{B2EAB1}{\textbf{\phantom{-}0.0276}} & -0.0554 \\\\  			
    \end{tabular}
    \caption{COMET scores with different SLT models (columns) and different audio segmentation methods (rows). Best result for \textit{automatic} segmentation scenario marked in bold and green.}
    \label{tab:comet}
    }
\end{table}

\subsection{Translation Quality with COMET}
\label{sec:comet}

For completeness, we present performance results measured with COMET \citep{rei-etal-2020-comet} in Table \ref{tab:comet}. This is evaluated outside of SLTev but we use the same resegmentation tool \citep{matusov-etal-2005-evaluating} to align the translations with the reference segments. The results show similar patterns as with BLEU and the context model paired with the merging window approach performs best among the automatic segmentation approaches on all language pairs. This approach even outperforms the gold segmentation on three language pairs. Note however that evaluating resegmented text with COMET may have some undesirable side-effects because the translated text is not always split at correct segmentation boundaries, e.g. the first token of a segment often is glued to the end of the previous segment.

We tested this with 200 gold segments for en-de and manually corrected the resegmentation of the original model output. While the BLEU score does not change much with these corrections (24.70 vs. 24.73), the COMET score jumps from -0.1128 to 0.0467 which is a larger improvement than some differences in Table \ref{tab:comet}. Since it is unclear if such resegmentation errors occur equally often in all our experiment setups, we only include the results with BLEU in the main body of the paper. We hypothesise that COMET has only seen well-formed sentences at training time and consequently is less reliable on such resegmented data. In the future, document-level neural evaluation metrics could be better suited for evaluating translations of unsegmented or automatically segmented audio in SLT.

\subsection{Flicker Results for Other Language Pairs}

\begin{figure*}[ht]
    \centering
    \includegraphics[width=0.9\textwidth]{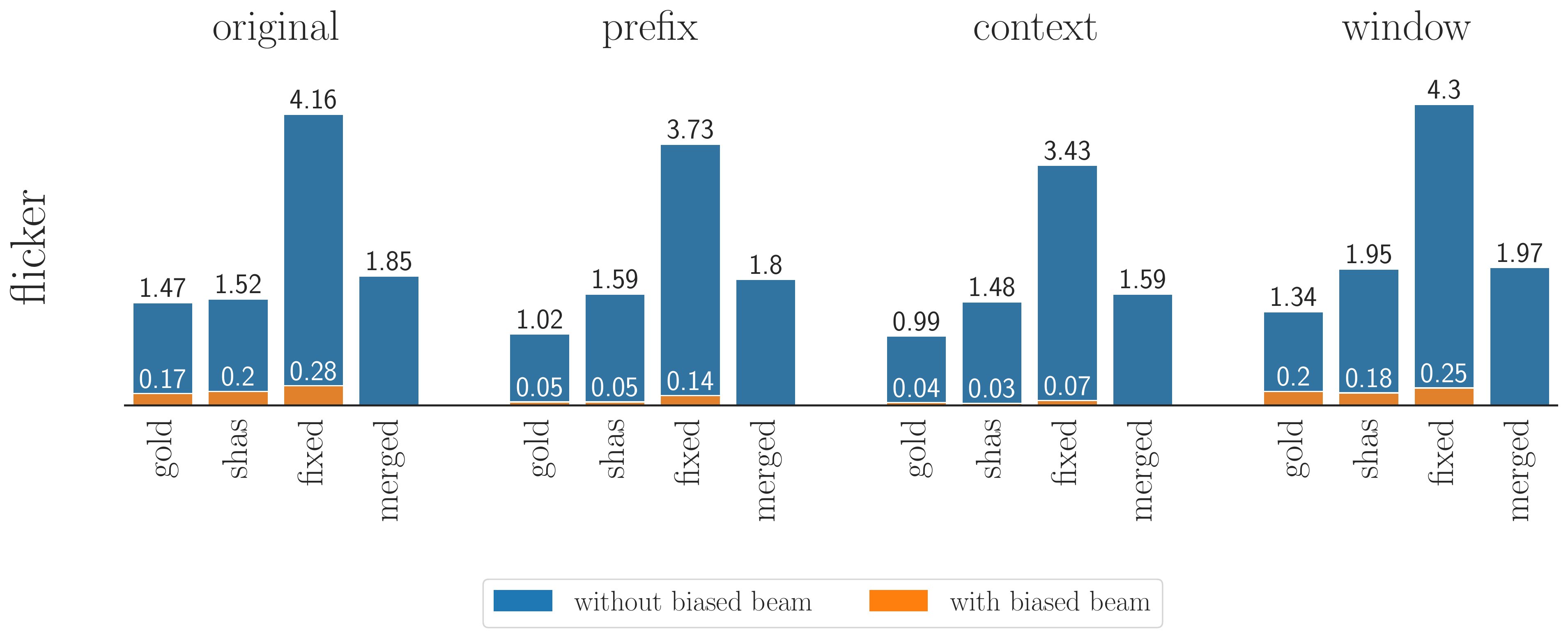}
    \caption{Flicker values for the different segmentation strategies and SLT models on the English-to-German test set.}
    \label{fig:flickerde}
\end{figure*}

\begin{figure*}[ht]
    \centering
    \includegraphics[width=0.9\textwidth]{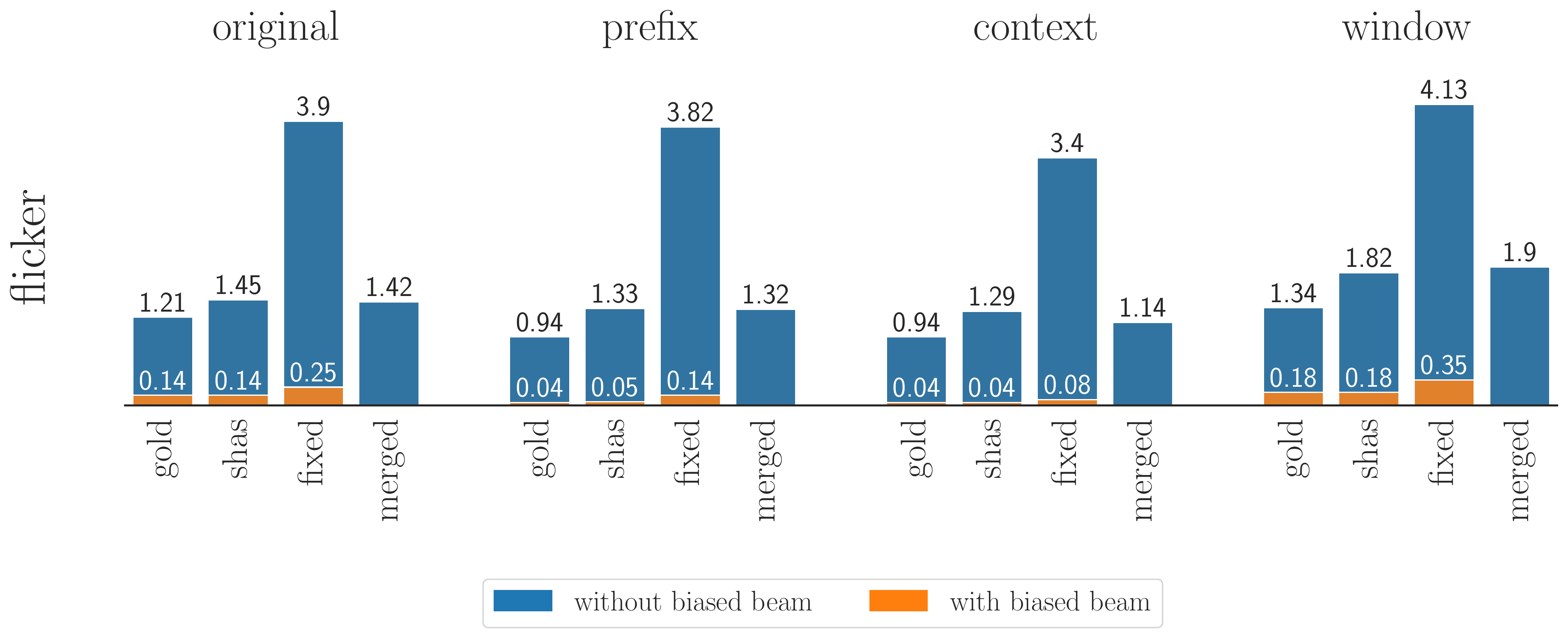}
    \caption{Flicker values for the different segmentation strategies and SLT models on the French-to-English test set.}
    \label{fig:flickerfr}
\end{figure*}

\begin{figure*}[ht]
    \centering
    \includegraphics[width=0.9\textwidth]{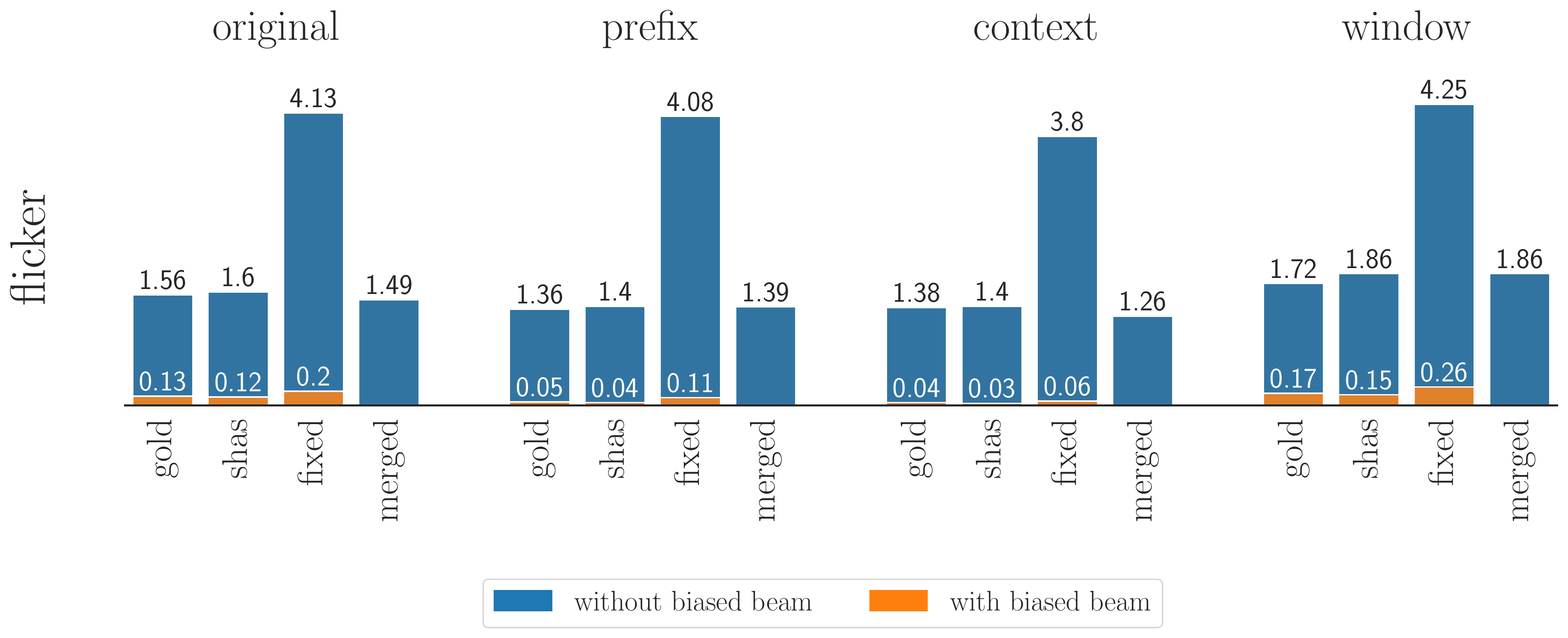}
    \caption{Flicker values for the different segmentation strategies and SLT models on the Italian-to-English test set.}
    \label{fig:flickerit}
\end{figure*}

\begin{figure*}[ht]
    \centering
    \includegraphics[width=0.9\textwidth]{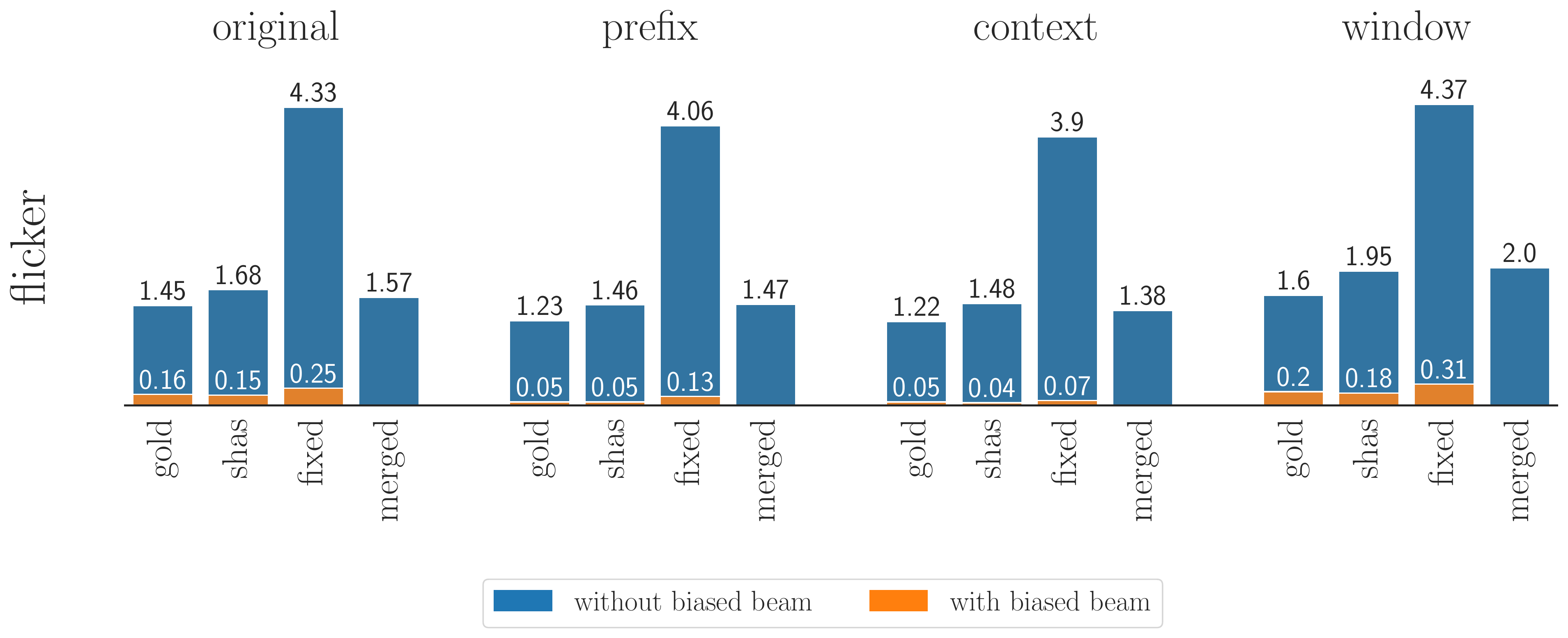}
    \caption{Flicker values for the different segmentation strategies and SLT models on the Portuguese-to-English test set.}
    \label{fig:flickerpt}
\end{figure*}

We present the same plots as in Section \ref{subsec:flicker} for English-to-German in Figure \ref{fig:flickerde}, French-to-English in Figure \ref{fig:flickerfr}, Italian-to-English in Figure \ref{fig:flickerit} and Portuguese-to-English in Figure \ref{fig:flickerpt}. The results follow the same patterns as the results for Spanish-English discussed in Section \ref{subsec:flicker}:

\begin{itemize}
    \item Fixed windows without biased beam search have the highest flicker.
    \item For the language pairs into English, the merging windows approach has lower flicker than SHAS if no biased beam search is used.
    \item Finetuning on context reduces flicker.
\end{itemize}

\subsection{Delay Results for Other Language Pairs}

\begin{figure*}[t]
    \centering
    \includegraphics[width=0.9\textwidth]{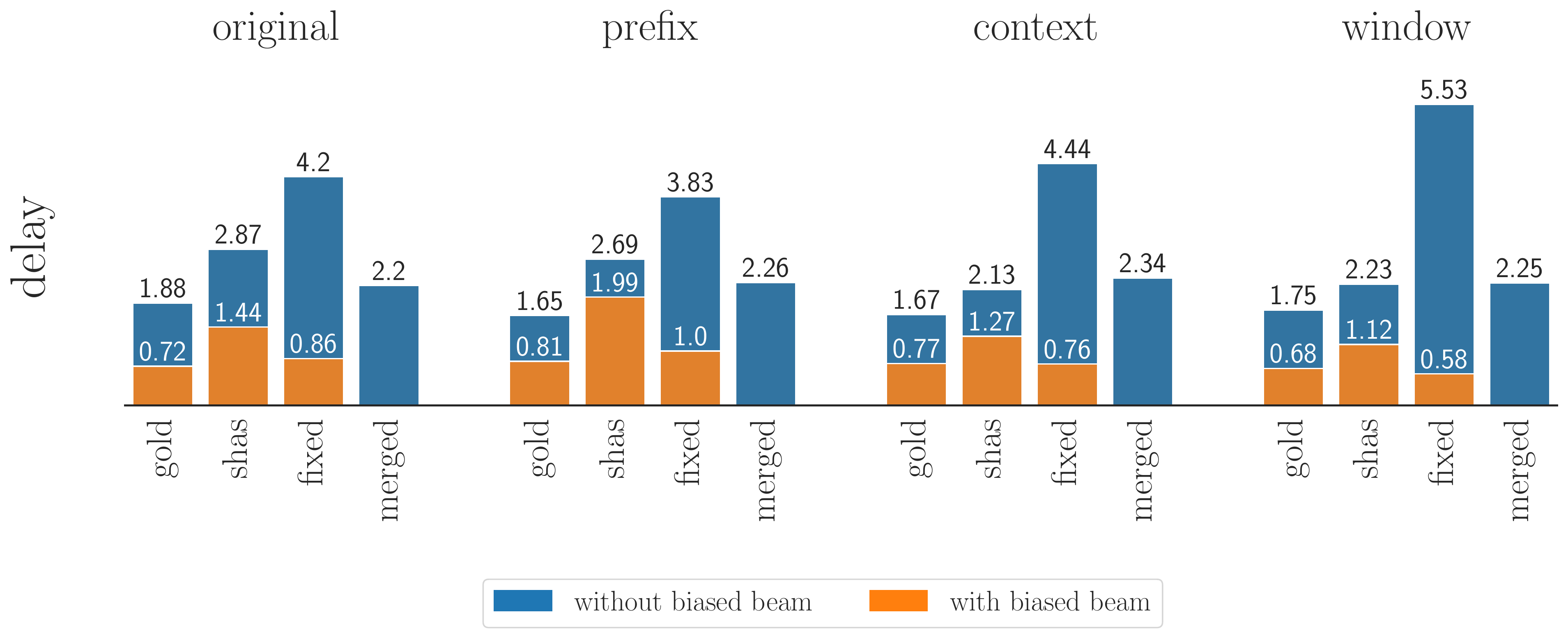}
    \caption{Delay values for the different segmentation strategies and SLT models on the English-to-German test set.}
    \label{fig:delayde}
\end{figure*}

\begin{figure*}[t]
    \centering
    \includegraphics[width=0.9\textwidth]{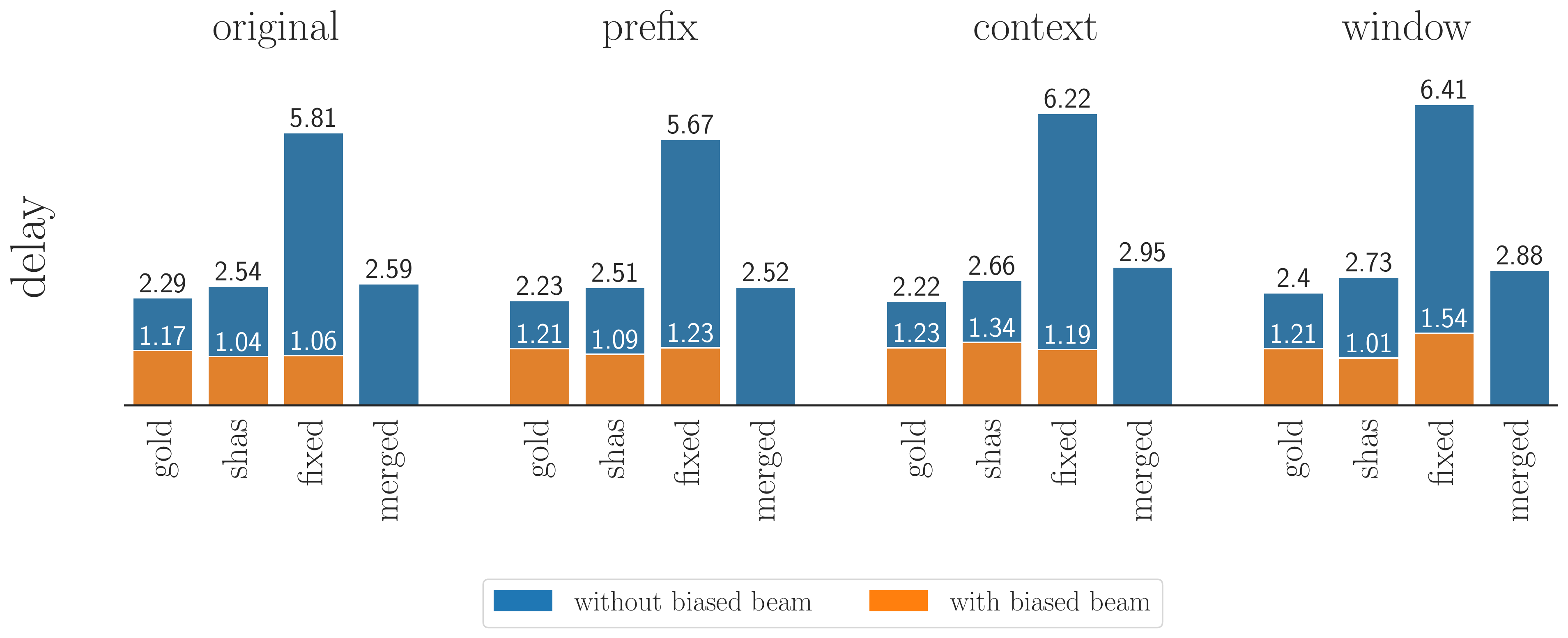}
    \caption{Delay values for the different segmentation strategies and SLT models on the French-to-English test set.}
    \label{fig:delayfr}
\end{figure*}

\begin{figure*}[t]
    \centering
    \includegraphics[width=0.9\textwidth]{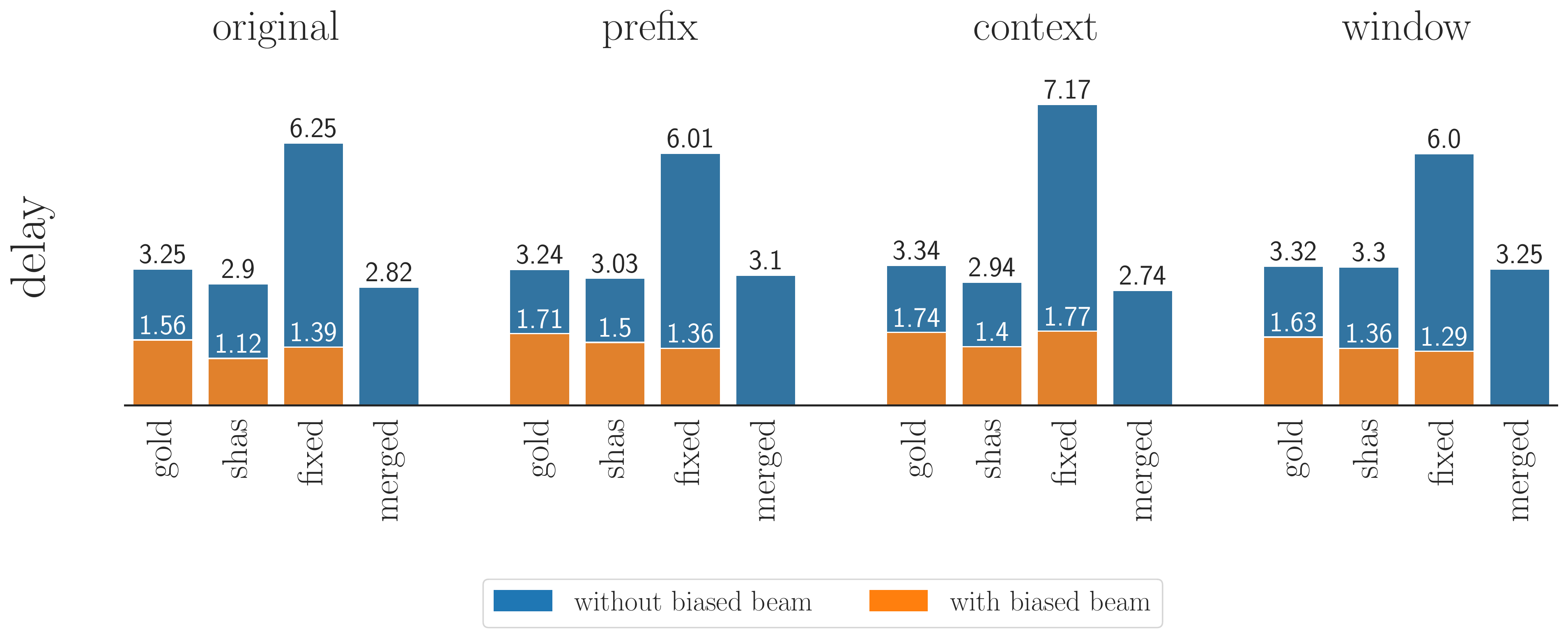}
    \caption{Delay values for the different segmentation strategies and SLT models on the Italian-to-English test set.}
    \label{fig:delayit}
\end{figure*}

\begin{figure*}[t]
    \centering
    \includegraphics[width=0.9\textwidth]{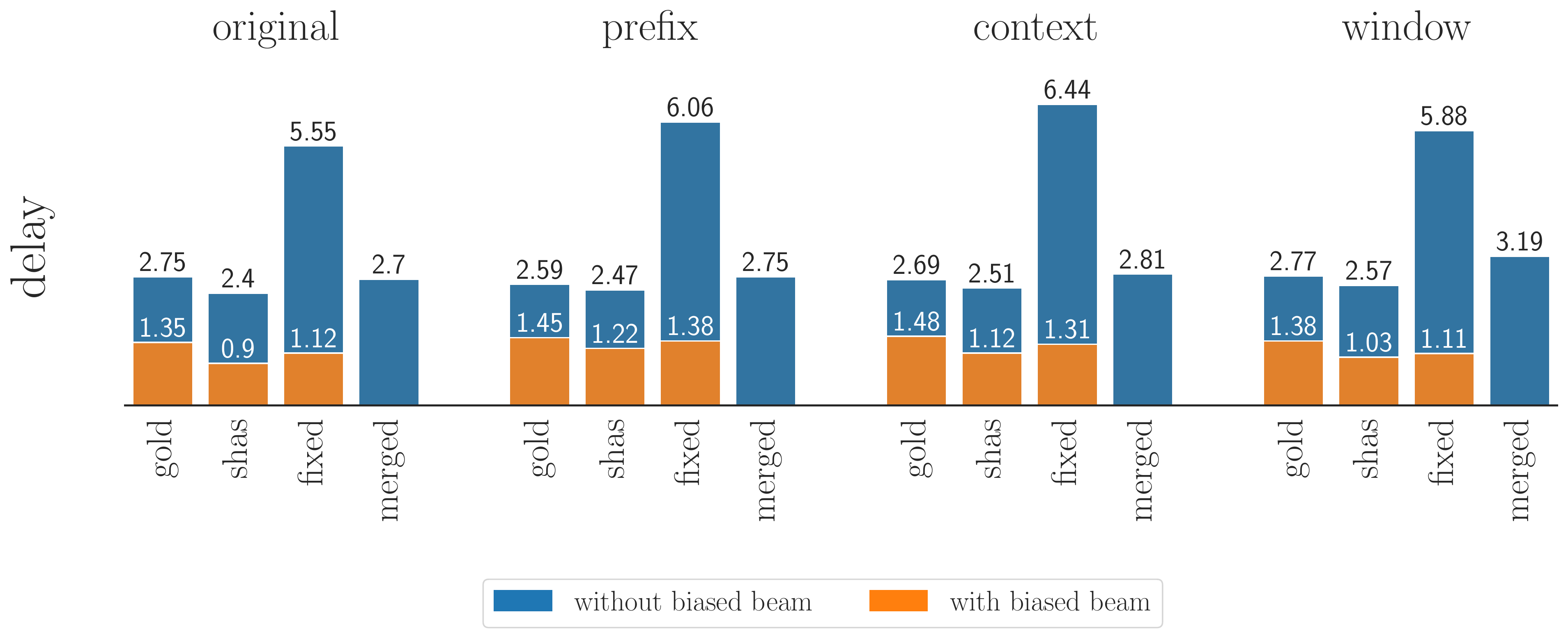}
    \caption{Delay values for the different segmentation strategies and SLT models on the Portuguese-to-English test set.}
    \label{fig:delaypt}
\end{figure*}

We present the same plots as in Section \ref{subsec:delay} for English-to-German in Figure \ref{fig:delayde}, French-to-English in Figure \ref{fig:delayfr}, Italian-to-English in Figure \ref{fig:delayit} and Portuguese-to-English in Figure \ref{fig:delaypt}. The results follow the same patterns as the results for Spanish-English discussed in Section \ref{subsec:delay}:

\begin{itemize}
    \item Fixed windows without biased beam search have the highest delay.
    \item The merging windows approach has comparable delay to SHAS if no biased beam search is used.
    \item Finetuning has less of an effect on delay than on flicker.
\end{itemize}

\end{document}